\RequirePackage{iftex}\ifPDFTeX\pdfoutput=1\fi  
\PassOptionsToPackage{hyphens}{url}  
\documentclass[]{ceurart}
\usepackage{booktabs}
\usepackage{etoolbox}
\AtBeginEnvironment{thebibliography}{\raggedright\interlinepenalty=10000\relax}

\begin{document}

\copyrightyear{2026}
\copyrightclause{Copyright for this paper by its authors.
  Use permitted under Creative Commons License Attribution 4.0 International (CC BY 4.0).}

\conference{GenAIECommerce'26: The Third Workshop on Agentic and
Generative AI for E-Commerce, co-located with RecSys, September 28, 2026,
Minneapolis, MN, USA}

\title{EvoRank: LLM-Guided Evolution of Multi-Objective Learning-to-Rank Pipelines}
\tnotemark[1]
\tnotetext[1]{Code, configurations, seeds, discovered programs, and all
machine-generated results are available at
\url{https://github.com/shabazpatel/evorank}.}

\author[1]{Rayhan Patel}[%
orcid=0009-0004-4882-2123,
email=rayhanbp@umd.edu,
]
\fnmark[1]
\author[2]{Shabaz Patel}[%
orcid=0000-0002-9617-7710,
email=shabaz.b.patel@gmail.com,
]
\cormark[1]
\fnmark[1]
\address[1]{University of Maryland, College Park, MD, USA}
\address[2]{Independent Researcher}
\cortext[1]{Corresponding author.}
\fntext[1]{These authors contributed equally.}

\begin{abstract}
We present EvoRank, an open autonomous ranking engineer: an LLM-guided
evolutionary loop that discovers complete Learning-to-Rank pipelines
(features, models, losses, ensembles) for multi-objective e-commerce
search. On the Expedia ICDM 2013 dataset, with
relevance, conversion, and revenue as competing objectives, three independent runs each converge within 50 iterations (about ten dollars) on interpretable pipelines
that beat an Optuna-tuned LambdaMART on 60k held-out queries, an advantage that persists at full data scale and places in the top 6
percent of the original competition. A first campaign, evolving only training objectives, builds the central
design rule: it appeared to work on its selection fold (the small dataset it uses to
pick winners) while a transfer audit, re-scoring winners on held-out data,
showed the gains were almost entirely fitness noise (the randomness of its
own scoring), and neither seeded domain knowledge nor richer diagnostic
feedback changed what transferred. The deciding quantity is
measurable in advance: search-space headroom relative to fitness noise. We
package this as a headroom gate that predicts, before any LLM spend, whether
the loop will pay off, and we release the system, the auditing tools, and a catalog of failure
modes with their guardrails, so teams can apply the procedure to their own ranking stacks.
\end{abstract}


\maketitle

\section{Introduction}
Automated discovery loops that pair a large language model with evolutionary
search over programs have produced striking results, and industrial systems
such as Meta's Ranking Engineer Agent apply the pattern to ranking at scale.
These systems are closed, and published results in the genre share a gap:
gains are overwhelmingly reported on the data the loop used to select
candidates. For ranking, where per-candidate evaluations are noisy by
necessity, this matters: selection under noise promotes lucky candidates,
and a loop can appear to make steady progress while discovering nothing
that transfers.

We contribute an open, fully audited account of when the pattern works
and when it silently does not, on a public e-commerce ranking task with
three competing objectives (relevance, conversion, revenue). Our contributions are:
(1) a quantified boundary condition for when such loops work: evolved training objectives whose selection-fold gains do not survive a
held-out transfer audit, because real effect sizes are smaller than the
noise of the loop's own scoring; (2) controlled ablations of guidance showing
that seeded domain knowledge and evidence-rich diagnostic feedback change the
speed and style of search but not what generalizes; (3) the positive case under identical auditing: pointing the same loop at a search space whose
effect sizes clear the noise floor yields transfer-audited pipelines,
better on every objective at once, in all three runs, which reconstruct and extend the ICDM 2013 challenge winners' hand-built
solutions; and (4) the released tool and its audit stack (the acceptance and stopping criteria applied to every result), including the search-space headroom gate, a cheap pre-check of whether a search space
will reward the loop.

\section{Related Work}
\textbf{LLM-guided program evolution.} FunSearch~\cite{funsearch} and
AlphaEvolve~\cite{alphaevolve} pair an LLM
mutation operator with evolutionary search over programs and automated
evaluators, with open engines~\cite{adaevolve,evox} reproducing the pattern;
industrial ranking agents (REA, GEARS)~\cite{rea} apply agentic
experimentation to production ranking behind closed doors. We add the missing measurement layer: the same loop shown selecting noise
in one search space and discovering in another, in an open, reproducible
instance of the pattern for tabular LTR.

\textbf{Automated loss discovery.} GLO~\cite{glo}, AM-LFS~\cite{amlfs},
AutoLoss-Zero~\cite{autolosszero}, and CSE-Autoloss~\cite{cseautoloss}
search loss functions from small mathematical building blocks, mostly for
vision; LambdaLoss~\cite{lambdaloss} is the hand-derived counterpart (a loss
built from a bound on the ranking metric);
RankEvolve~\cite{rankevolve} applies evolutionary search to unsupervised
lexical retrieval with a single-number score. We bring loss search to supervised multi-objective ranking with
gradient-boosted trees (GBDT) via LLM mutations over full gradient code, and map a boundary this literature has
not measured: the family's headroom sits below the noise floor of any
evaluation fast enough to run inside a loop, so discovered losses do not
transfer.

\textbf{Reflective feedback for LLM optimization.} Reflexion~\cite{reflexion},
ProTeGi~\cite{protegi}, OPRO~\cite{opro}, and GEPA~\cite{gepa} show that richer textual feedback (traces, textual gradients)
improves LLM-driven search. We measure that assumption's boundary: feedback that inspects the
candidate's own training gradients (possible because the evolved program is
a loss function) changed nothing about what transferred across three runs,
while simply enlarging the evaluation fold tripled it.

\textbf{Agentic AutoML and LLM feature engineering.} CAAFE~\cite{caafe}
generates features for tabular classification; AIDE~\cite{aide},
ML-Agent~\cite{mlagent}, and related agents iterate over ML pipelines, typically evaluated by single-objective validation
accuracy. We differ in the ranking domain (query-grouped data, per-query features
and ensembling), in transfer-audited selection as the acceptance bar, and
in a quantified catalog of target-encoding leakage modes.

\textbf{LTR foundations and the benchmark.} LambdaMART~\cite{lambdamart}
remains the standard for feature-based ranking, where gradient-boosted trees
still match or beat neural rankers~\cite{qin2021}; the ICDM 2013 challenge
solutions~\cite{liu2013} document the winning feature recipes and score-blending ensembles; a 2024 survey~\cite{survey2024} notes that public
datasets with monetary outcome labels remain scarce, anchoring open
multi-objective work to this benchmark. Our loop rediscovers and extends the essential structure of those
solutions autonomously, for about ten dollars.

\section{System and Audit Methodology}

\begin{figure}
  \centering
  \includegraphics[width=\linewidth]{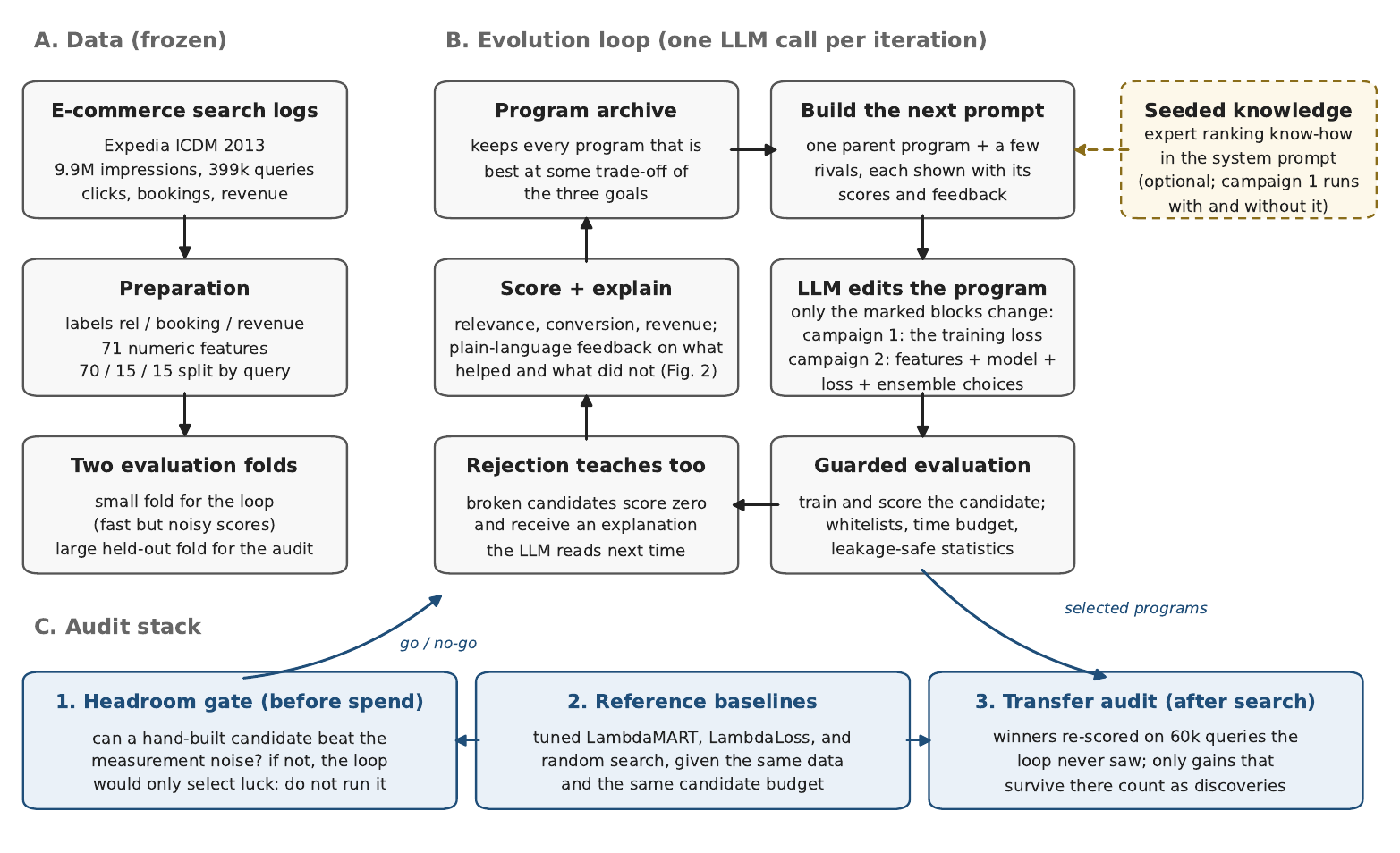}
  \caption{The EvoRank system. Lane A: frozen data preparation producing a
  small scoring fold and a large held-out fold. Lane B: the loop, one LLM call per iteration, from archive through
  prompt, edit, guarded evaluation, and feedback (Figure~\ref{fig:feedback}) back to the
  archive. Dashed: optional expert knowledge, ablated in campaign 1. Lane C: the audit stack: pre-spend headroom gate, baselines on
  equal terms, and the held-out audit that decides what counts.}
  \label{fig:system}
\end{figure}

\subsection{Background: the evolutionary engine}
EvoRank builds on SkyDiscover, an open framework for LLM-guided evolutionary
program search in the AlphaEvolve family~\cite{alphaevolve,adaevolve}
(Figure~\ref{fig:system}). The unit of evolution is a source
file in which mutable regions are delimited by EVOLVE-BLOCK markers;
everything outside the markers is frozen. Each iteration samples a parent (plus context programs) from a
program database, assembles a prompt with their source, scores, and
feedback, makes one LLM call that edits the mutable regions, evaluates the
candidate with a user-supplied \texttt{evaluate} function, and inserts it
back into the database.

We use SkyDiscover's AdaEvolve search controller~\cite{adaevolve}. Its program database keeps several semi-isolated subpopulations
(islands; two here, spawning up to five on stagnation) that periodically exchange their best programs; a bandit allocates
iterations to islands by recent improvement, and each island adapts its
exploration intensity accordingly. The database keeps a Pareto archive, the set of best available trade-offs
over the three objectives (ndcg, booking ndcg, pooled dollar-weighted
revenue), with a single-number score used only for tie-breaking.

\subsection{Preliminaries: the algorithms in brief}
\emph{Ranking and NDCG@k.} A query returns a list of items with relevance
labels. NDCG@k scores how well a model orders each list: relevant items near the
top earn gain discounted by position, normalized so 1.0 is perfect. We average it per query (for revenue
we instead pool dollars, so high-value queries weigh more).

\emph{GBDT and LambdaMART.} Gradient-boosted decision trees
(XGBoost~\cite{xgboost}, LightGBM~\cite{lightgbm}) fit a sequence of small trees, each correcting the previous ones via
gradients of a loss. LambdaMART is the ranking instance:
for every within-query pair where one item outranks another in the
labels, it applies a pairwise gradient scaled by how much swapping the two
would change NDCG, concentrating training pressure where the metric is at
stake. XGBoost and LightGBM implement it; campaign 1 evolves exactly this
gradient function.

\emph{Hyperparameter tuning with Optuna.} A GBDT's quality depends on
learning rate, depth, subsampling, and regularization. Optuna~\cite{optuna}
searches that space with Bayesian trials scored on validation data. Baselines get 40 trials, re-run at every training-set size we report,
since optimal settings change with data volume.

\emph{Ensembling by per-query z-scores.} Different model families make
partially independent errors; because their scores live on different scales,
each model's scores are standardized within each query before a weighted
sum. It was the challenge winners' choice and what our pipelines select.

\emph{Out-of-fold target encoding.} Encoding a categorical column by the average label of its value (a
hotel's historical booking rate) leaks labels into features unless each row
is encoded from other query-folds only.

\emph{The statistics.} A paired query bootstrap resamples queries and recomputes the difference
between two systems many times (10,000 in the audit, a few hundred inside
the loop's noise gate), treating the query as the unit of noise. Pareto dominance
means at least as good on every objective and better on one; hypervolume
measures how much objective space the non-dominated set covers.

\subsection{Task, data, and evaluators}
Data is the public Expedia Personalized Hotel Search training file (9.9M
impressions, 399k queries), split 70/15/15 by query. Each shown hotel gets a
graded relevance label: 5 if booked, 1 if clicked, 0 otherwise. The loop
evaluates candidates on a small fitness fold of 8k training queries: a candidate trains a gradient-boosted ranker under a fixed budget and is
scored on the fold's held-back validation queries; that score is its
fitness, and the small fold keeps each evaluation between 5 and 40 seconds. Some
candidates are broken rather than weak, for example gradients that turn
infinite, training that exceeds the time budget, or a malformed pipeline
specification. These receive zero fitness together with a message explaining
what went wrong, which enters the next prompt, so the guardrails double as
teaching signals.

\subsection{The audit stack: acceptance and stopping criteria for the loop}
\label{sec:audit}
The audit stack, applied identically to every campaign: (a) the headroom
gate, run before any LLM spend: one hand-built candidate must beat the
fitness fold's noise floor by a set margin, or the search space is not
worth searching; (b) paired query
bootstrap for every comparison, with noise-gated deltas inside the loop's
own feedback; (c) equal-budget random-search controls; (d) equal tuning on both sides (discovered programs get the same Optuna
budget as the baseline); (e) the transfer audit itself, re-scoring every
checkpoint's selected program on 60k held-out queries; (f) three-objective
frontier quality (hypervolume); and (g) component-removal tests on
discovered programs. The tuned baseline throughout is LambdaMART (XGBoost
rank:ndcg) with 40 Optuna trials.

\section{Campaign 1: Objective Space, Where the Loop Selects Noise}
The evolvable piece is the LambdaMART gradient function, computed per
query group (the set of hotels returned for one search), with access to each impression's raw behavioral signals (click, booking,
revenue for each shown hotel); training configuration, features, and
metrics are frozen, and the starting version reproduces XGBoost's built-in
rank:ndcg within noise.

On its selection fold, evolved objectives beat the default baseline and
improve steadily. The transfer audit reverses the picture: final selections gain $+0.0004$
to $+0.0014$ NDCG@10 on held-out queries, the best evolved objective beats
its own starting program by $+0.0002$ (two-sided $p=0.76$), an equal-budget random
search transfers as well or better ($p=0.03$ in its favor), and every method
is significantly below the tuned baseline ($p<0.001$). In plain terms: on
unseen data the evolved objectives are indistinguishable from where they
started, blind mutation did as well as guided search, and ordinary tuning
beat both. The mechanism is measured: component-removal tests put a single
component's effect at 0.0005 to 0.008 NDCG@10, most below 0.005, while the
selection fold's measurement noise (95 percent interval half-width) is
$\pm 0.007$ to $0.009$ NDCG@10 and wider on booking and revenue, so
picking the best score means picking luck. Enlarging the fitness fold's validation side from 1.6k to 6k queries
halves that noise and triples the gain surviving transfer, the only
intervention that moved it.

Guidance ablations: seeding the prompt with curated domain knowledge makes
the search faster, not truer. Two of three seeded runs need only 1 to 3
iterations to match the best score an equal-budget random search reaches in
its entire run (the third never does); unseeded runs need 7 to 23. Seeded runs also end with a better three-objective frontier in every run. What transfers
to held-out data is unchanged. A diagnostic feedback channel (noise-gated deltas, per-segment breakdowns,
probes into the candidate's training gradients) changes nothing measurable across three runs.

Analysis beat further search: removing the best objective's components one
at a time (five evaluations in total) explained its behavior and exposed one
component, a blend weight that adapted per query, as actively harmful;
fixing that weight to a constant improved all three metrics. The family also
flattens at scale: trained on the full dataset, every objective, including
default rank:ndcg, lands within 0.001 NDCG@10 of every other.

\section{Campaign 2: Pipeline Space, Where the Loop Discovers}
Campaign 2 widens the search from the loss alone to the whole pipeline.
The LLM now edits two blocks: feature-construction code over semantically
named columns, where any statistic touching labels must come from helper
functions with leakage guards built in (counts, out-of-fold rates,
quantiles, and a train-fold linear score over query-level columns), and a pipeline specification from a whitelist (one to three members from XGBoost, LightGBM, or sklearn; ranking,
pointwise, binary, or campaign-1 losses; z-score or rank-mean blending). Feedback is stage-attributed (each pipeline stage gets its own line) so
the loop can tell which axis earned each change; Figure~\ref{fig:feedback} shows an excerpt.

\begin{figure}
\footnotesize
\begin{verbatim}
scores: ndcg=0.4357 (+0.0139 vs seed, +/-0.0054, SIGNIFICANT) |
  book_ndcg=0.4604 (+0.0142, SIGNIFICANT) | revenue=0.4309 (+0.0217, SIGNIFICANT)
members: [0] xgb/rank:ndcg: val ndcg 0.4325 (2s)
         [1] lgbm/lambdarank: val ndcg 0.4261 (4s)
         [2] sklearn/ert: val ndcg 0.4171 (28s)
ensemble: zscore_weighted [0.55, 0.33, 0.12] = 0.4357,
  best single 0.4325 (+0.0033)
\end{verbatim}
\caption{Stage-attributed feedback for one campaign 2 candidate, excerpted
from the run logs (feature-stage line elided). The +/- changes are measured
against the starting pipeline and flagged SIGNIFICANT only above noise;
per-member lines attribute quality to model and loss choices; the ensemble
line reports the margin over the best member.}
\label{fig:feedback}
\end{figure}

Before any LLM spend, we applied the headroom gate (Section~\ref{sec:audit}): a hand-built candidate
from the ICDM 2013 literature had to beat the noise floor by 2.5 times, and
within-query rank features plus count encodings did ($+0.013$ NDCG@10).
Building that candidate surfaced three successively subtler leakage bugs.
Encoding a hotel by its booking rate on the training fold costs $-0.10$
NDCG: the model memorizes its own labels. The textbook fix, leaving each
row's own label out of its encoding, is worse at $-0.13$: the exclusion
shifts the encoded value in a direction that depends on the label, and the
trees read that shift as the label itself. Only out-of-fold encodings, with
folds split by query, are clean. The audit caught each bug before any
search depended on it.

\begin{table}
  \caption{Campaign 2 transfer audit: trained on the fitness fold, tested on
  59{,}902 held-out queries. All runs beat the tuned baseline on NDCG@10; the
  best run (s2) beats it on all three objectives. Bold: best per column.}
  \label{tab:transfer}
  \begin{tabular}{lcccc}
    \toprule
    Method & Val NDCG@10 & Test NDCG@10 & Test book & Test revenue \\
    \midrule
    Starting pipeline & 0.4218 & 0.4222 & 0.4443 & 0.4105 \\
    LambdaMART + Optuna & 0.4330 & 0.4296 & 0.4527 & 0.4231 \\
    Pipeline s0 & 0.4356 & 0.4316 & 0.4545 & \textbf{0.4257} \\
    Pipeline s1 & 0.4359 & 0.4322 & 0.4553 & 0.4215 \\
    Pipeline s2 & 0.4398 & \textbf{0.4352} & \textbf{0.4582} & 0.4239 \\
    \bottomrule
  \end{tabular}
\end{table}

What the search found (Figure~\ref{fig:discovered}): three runs, at 50
iterations and ten dollars each, converged on the same interpretable core: within-query percentile
ranks of price, star rating, review and location scores, count encodings of
property and destination identifiers, a lean feature set (the programs' own comments cite the earlier feedback
showing that more features hurt), and a three-member ensemble of unlike
models blended by per-query z-scores. One provenance caveat: the seeded knowledge included the gate's finding
that rank and count features carry headroom, so their presence is partly
prompted; the specific constructions, model, loss, ensemble, and weight
choices, and unprompted inventions (per-query price normalization,
percentile-filling a mostly-missing location score) are the search's own.

\begin{figure}
\footnotesize
\begin{verbatim}
# discovered by run s1 (excerpt; ellipses and "# 5 columns" are editorial)
for c in ["price_usd", "prop_starrating", ...]:   # 5 columns
    out[f"{c}_qrank"] = df.groupby("qid")[c].rank(pct=True)
for c in ["prop_id", "srch_destination_id"]:
    out[f"{c}_count"] = stats.count(df, c)

PIPELINE = {"models": [
    {"family": "xgb",     "objective": "rank:ndcg",  ...},
    {"family": "lgbm",    "objective": "lambdarank", ...},
    {"family": "sklearn", "objective": "ert",        ...}],
  "ensemble": {"type": "zscore_weighted", "weights": [0.5, 0.35, 0.15]}}
\end{verbatim}
\caption{The discovered program is readable code: feature excerpt and
pipeline specification from run s1 (full programs in the repository).}
\label{fig:discovered}
\end{figure}

Does it transfer? Yes (Table~\ref{tab:transfer}): roughly 70 percent of
selection-fold gains survive on 60k held-out queries, and all three runs
beat the tuned baseline there. A paired query bootstrap (10,000 resamples) makes
this formal: the best run wins by +0.0057 NDCG@10 (95 percent CI [0.0041,
0.0073], one-sided p < 0.0001) and +0.0056 booking NDCG@10 (p < 0.0001),
with revenue positive but within noise (+0.0008, CI [-0.0034, 0.0047]); the
full-scale comparison shows the same pattern (+0.0075 NDCG@10, p < 0.0001).
The multi-objective claim is therefore precise: significantly better
relevance and conversion at matching revenue.

\begin{table}
  \caption{Full-scale comparison (280k train, 60k test queries), including
  NDCG@38, the challenge metric. The last row gives the pipeline's XGBoost
  member the baseline's full-scale tuned settings, scales its LightGBM
  member to match (255 leaves, 300 trees), and drops the third member; it
  had no search or tuning of its own.}
  \label{tab:fullscale}
  \begin{tabular}{lcccc}
    \toprule
    Method & NDCG@10 & NDCG@38 & Book@10 & Revenue@10 \\
    \midrule
    Tuned baseline (small-fold tuning) & 0.4349 & 0.4983 & 0.4588 & 0.4300 \\
    Tuned baseline (full-scale tuning) & 0.4544 & 0.5132 & 0.4805 & 0.4565 \\
    Pipeline s1 (small-fold tuning) & 0.4490 & 0.5080 & 0.4738 & 0.4483 \\
    Pipeline s1 (baseline's tuned settings, 2 members) & \textbf{0.4619} & \textbf{0.5185} & \textbf{0.4883} & \textbf{0.4585} \\
    \bottomrule
  \end{tabular}
\end{table}

Full-dataset training yields its own methodology lesson
(Table~\ref{tab:fullscale}): against a baseline tuned on the small fold the
pipelines win easily; re-tuning the baseline at full scale reverses that
verdict; and only a comparison with comparable capacity on both sides
isolates the discovered structure, which then wins on all four numbers. The
margin splits across both components: the discovered features alone, under
the same tuned single model, reach 0.4587 NDCG@10 (+0.0043); the two-member
ensemble adds +0.0032 to 0.4619.
We also scored both final models on the original competition's hidden
test set
via Kaggle late submission (one shot each, no leaderboard iteration): the
discovered pipeline reaches 0.5196 private NDCG@38, which would have placed
20th of 340 teams (top 6 percent) in the 2013 challenge, with the tuned
baseline at 0.5140 (30th), so ten leaderboard places come from the discovered structure. The winning hand-built ensembles remain ahead (0.5398; our cited solution
reports 0.5310), and our submission benefits from hindsight, since the
winners' published lessons are in the seeded knowledge.

\section{Discussion}

\begin{figure}
  \centering
  \includegraphics[width=0.92\linewidth]{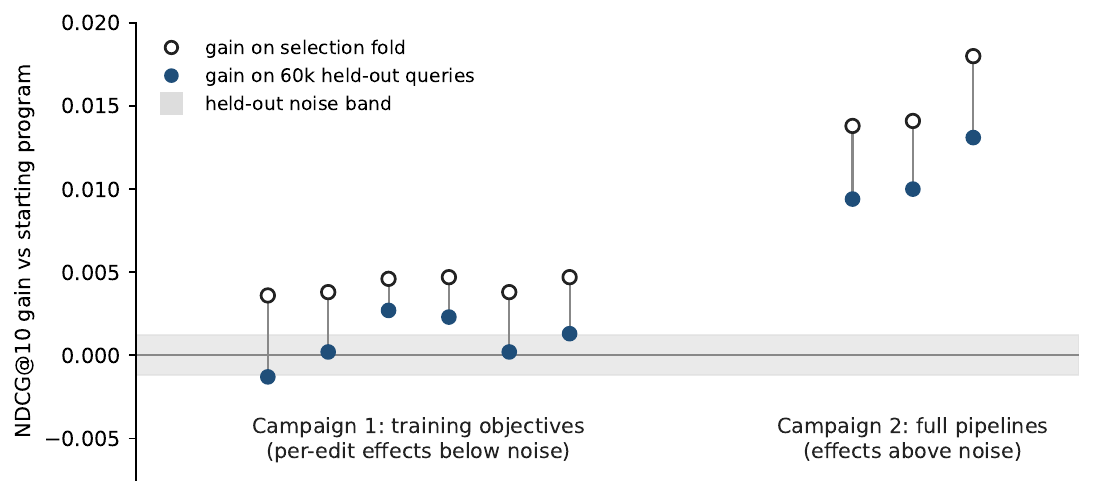}
  \caption{The study in one picture: per-run NDCG@10 gains over the
  starting program, on the selection fold (open circles) and on 60k held-out queries
  (filled). Campaign 1's gains collapse into the noise band; campaign 2's
  survive.}
  \label{fig:contrast}
\end{figure}

The two campaigns differ in exactly one variable, the search space, and the
same loop produced noise in one and transfer-audited Pareto dominance in the
other (Figure~\ref{fig:contrast}). The deciding quantity is measurable in advance: the ratio of
achievable effect sizes to fitness-fold noise. We suggest the headroom gate as a cheap pre-spend go/no-go check for
discovery loops, and transfer-audited selection as the reporting standard. Our results also rank where a team should invest next: evaluators first
(everything that moved outcomes lived there: fold size, leakage-safe
statistics, rejection messages the model learns from, the gate, and the
audit); application spaces second (headroom relative to noise decides the
outcome, so choosing the space is choosing the result); search algorithms
last (a recalibrated AdaEvolve and an EvoX meta-evolution
variant~\cite{evox}, each at double budget, landed inside the spread of the
original three runs, their internal signals drowned by the same noise). The
natural next step is multi-fidelity evaluation: screen on the cheap fold,
promote survivors to full scale before selection, and so expose the
scale-dependent features (per-entity rates, position effects) behind much
of our gap to the 2013 winners.

\textbf{Limitations and generalization.} Our evidence comes from one
dataset, now thirteen years old, because newer public LTR datasets lack the
conversion and revenue labels a multi-objective study needs~\cite{survey2024}.
The audit stack itself is dataset-agnostic: it needs only a small fitness
fold, a larger held-out fold, a paired bootstrap on the metric of interest,
and one strong tuned baseline, and the headroom gate is a recipe for any
query-grouped ranking or recommendation task: measure the fitness fold's
noise by query bootstrap, hand-build one candidate from the domain's
literature, and require its gain to clear that noise by a set margin (2.5
times here). We compared two search spaces; intermediate ones (features
plus loss, model plus loss) are untested, and the gate is the tool for
ranking them before spending. Three runs per condition suffice for the
paired-bootstrap headline, which every run cleared, but not to
characterize the variance of the search itself. We did not compare against other LLM-guided discovery
engines~\cite{alphaevolve,gepa}; the audit is engine-agnostic (the EvoX arm
above is one such swap), and a controlled comparison is future work.
Finally, revenue is a proxy metric, and discovered pipelines were audited
for transfer, not for online effects.

\section{Conclusion}
For teams bringing LLM-driven discovery to their own ranking systems, this
study distills to a working procedure: measure the noise of the evaluation
you can afford; verify with one hand-built candidate that your search space
offers effects above that noise; only then run the loop, with leakage-safe
statistics and rejection messages the model can learn from; and let a
held-out audit, not the loop's own scores, decide what counts. Code, configs, seeds,
logs, discovered programs, and audit scripts are released at
\url{https://github.com/shabazpatel/evorank}.

\end{document}